\def\year{2020}\relax
\documentclass[letterpaper]{article} 
\usepackage{aaai20}  
\usepackage{times}  
\usepackage{helvet} 
\usepackage{courier}  
\usepackage[hyphens]{url}  
\usepackage{graphicx} 
\usepackage{amsmath}
\usepackage{amsfonts}
\usepackage{amsthm}
\usepackage[english]{babel}
\usepackage{graphicx}  
\newtheorem{theorem}{Theorem}[section]

\title{FairyTED: A Fair Rating Predictor for TED Talk Data}
 \author{Rupam Acharyya,\textsuperscript{\rm $\dagger$}
Shouman Das,\textsuperscript{\rm $\dagger$}
Ankani Chattoraj,\textsuperscript{\rm $\dagger$}
Md. Iftekhar Tanveer\textsuperscript{\rm *}\\
\textsuperscript{\rm $\dagger$}University of Rochester,
\textsuperscript{\rm *}Comcast Applied AI Research\\
racharyy@cs.rochester.edu, shouman.das@rochester.edu,\\ achattor@ur.rochester.edu,
mdiftekhar\_tanveer@comcast.com 
}

\begin{document}

\maketitle

\begin{abstract}
With the recent trend of applying machine learning in every aspect of human life, it is important to incorporate fairness into the core of the predictive algorithms. We address the problem of predicting the quality of public speeches while being fair with respect to sensitive attributes of the speakers, e.g. \emph{gender} and \emph{race}. We use the TED talks as an input repository of public speeches because it consists of speakers from a diverse community and has a wide outreach. Utilizing the theories of \emph{Causal Models}, \emph{Counterfactual Fairness} and state-of-the-art neural language models, we propose a mathematical framework for fair prediction of the public speaking quality. We employ grounded assumptions to construct a causal model capturing how different \emph{attributes} affect public speaking quality. This causal model contributes in generating counterfactual data to train a \emph{fair} predictive model. Our framework is general enough to utilize any assumption within the causal model. Experimental results show that while prediction accuracy is comparable to recent work on this dataset, our predictions are counterfactually fair with respect to a novel metric when compared to true data labels. The FairyTED setup not only allows organizers to make informed and diverse selection of speakers from the unobserved counterfactual possibilities but it also ensures that viewers and new users are not influenced by unfair and unbalanced ratings from arbitrary visitors to the \url{ted.com} website when deciding to view a talk.
\end{abstract}

\section{Introduction}\label{introduction}
In recent times, artificial intelligence is being used in consequential decision making.  Governments make use of it in criminal justice system to predict recidivism~\cite{brennan2009evaluating,tollenaar2013method} which affects the decision about bail, sentencing and parole. Various firms are also using machine learning algorithms to examine and filter resumes of job applicants~\cite{nguyen2016hirability,chen2017automated,naim2016automated} which is crucial for the growth of a company. Machine learning algorithms are also being used to evaluate human's social skills such as presentation performance~\cite{Chen2017a,Tanveer2015}, essay grading~\cite{alikaniotis2016automatic,taghipour2016neural} etc.

To solve such decision making problems, machine learning algorithms are trained on massive datasets that are usually collected in the wild. Due to difficulties in the manual curation or adjustment over large dataset, it is likely that the data capture unwanted bias towards the underrepresented group based on race, gender or ethnicity. Such bias results in unfair decision making systems, leading to unwanted and often catastrophic consequences to human life and society. For example, the recognition rates of pedestrians in autonomous vehicles are reported to be not equally accurate for all groups of people~\cite{wilson2019predictive}. Matthew et al.~\cite{kay2015unequal} showed that societal bias gets reflected in the machine learning algorithms through biased dataset and causes representational harm for occupations. Face recognition has been found to be not as effective for people with different skin tones. Dark-skinned females have $43$ times higher detection error than light-skinned males ~\cite{buolamwini2018gender}.

In this work, we propose a predictive framework that tackles the issue of designing a fair prediction system from biased data. As an application scenario, we choose the problem of fair rating prediction in the TED talks. TED talks cover a wide variety of topics and influence audience by educating and inspiring them. In addition, it consists of speakers from a diverse community with imbalances in the age, gender and ethnic attributes. The ratings are provided by spontaneous visitors to the TED talk website. A machine learning algorithm trained solely from the audience ratings will have a possibility of the predicted rating being biased by sensitive attributes of the speakers. 

It is a challenging problem because human behavior is driven by numerous factors and hence have huge variability. It is difficult to know the way these factors interact among each other. In addition, uncovering the true interaction model may not be feasible and often expensive. Even though the sharing platforms such as YouTube, Massive Open Online Courses (MOOC), or \url{ted.com} make it possible to collect a large amount of observational data, these platforms do not correct for bias and unfair ratings.

In this work, we utilize \emph{causal models}~\cite{pearl2009causal} to define possible dependencies between attributes of the data. We then address the issue of not knowing the true interaction model by averaging outputs of predictors across several possible causes. Further using these causal models we generate \emph{counterfactual samples} of the sensitive attributes. These counterfactual samples are the key components in our fair prediction framework (adapted from ~\citeauthor{kusner2017counterfactual}~\shortcite{russell2017worlds}) and help reducing bias in the ratings with respect to sensitive attributes. Finally, we introduce a novel metric to quantify the degree of fairness employed by our FairyTED pipeline. To the best of our knowledge, FairyTED is the first fair prediction pipeline for public speaking dataset and can be applied to any dataset of similar grounds. Apart from the theoretical contribution, our work also has practical implications in helping both the viewers and the organizers make informed and unbiased choices for selection of talks and speakers.

\section{Related Works}\label{relatedworks}
There has been a rising interest in developing fair algorithms focused to mitigate the bias arising from discriminatory preferences of attributes such as gender, ethnicity, race, etc. These can cause bias across various domain, from college admissions process to criminal justice \cite{bickel1975sex,brennan2009evaluating}.  Training a machine learning algorithm with an objective of getting higher prediction accuracy can be sometimes unfair towards underrepresented groups in the dataset. For example, it has been shown in ~\cite{bolukbasi2016man} that the geometry of word embedding trained with traditional machine learning algorithms reflect gender stereotypes present in our society. Since machine learning models are used to take important and sensitive decisions including credit score prediction, loan applications assessment or predicting crime scenes, a careful approach should be designed to make traditional predictive model fair.~\citeauthor{baeza2018bias} emphasized the importance of increasing awareness about fairness in web based system. ~\citeauthor{calders2010three}~\shortcite{calders2010three} proposed methods for designing discrimination-free bayesian classifier. \citeauthor{dwork2012fairness}~\shortcite{dwork2012fairness} formulated fairness as an optimization problem and made use of a task specific similarity metric which describes the similarity of two individuals for the classification task. \citeauthor{grgic2016case}~\shortcite{grgic2016case} defined the notion of process fairness by focusing on the process of decision making rather than outcome of the classifier. \citeauthor{schumann2019transfer}~\shortcite{schumann2019transfer} proposed a framework with theoretical gurantees to transfer fairness in machine learning across various domains. Other related research has been done where the main focus is to quantify the unfairness in a machine learning algorithms and create a model for a certain dataset. Readers are referred to
~\cite{kamiran2009classifying,kamishima2011fairness,joseph2016rawlsian,garg2019counterfactual}.
For a recent complete survey see~\cite{mehrabi2019survey}.
The notion of fairness in a prediction algorithm has been defined in various ways based on researcher's assumption of fairness ~\cite{zliobaite2015survey,zafar2017fairness} . We follow the causal approach first introduced by ~\citeauthor{kusner2017counterfactual}~\shortcite{kusner2017counterfactual} to address the notion of fairness in a machine learning model. For the causal framework, we have adopted the definition of \cite{pearl2009causal}.

\section{Preliminaries}\label{prelims}
\subsection{Causal Model Definition}
Following general convention we define causal model as a Directed Acyclic Graph (DAG) with a set of nodes ($N$) and edges $(E \subseteq N \times N)$. Let $Pa_i = \{ n_j| (n_j,n_i) \in E\}$ denote the set of parents of node $n_i$. Adapting the conventions used in \citeauthor{pearl2009causal}~\shortcite{pearl2009causal}, \citeauthor{kusner2017counterfactual}~\shortcite{kusner2017counterfactual}, we then define the main characteristics of the causal DAGs. Each causal DAG consists of the triple ($\mathcal{U}$, $\mathcal{V}, \mathcal{F}$) where,
\begin{itemize}
    \item $\mathcal{U}$ denotes the set of unobserved variables in the outside world that influence observed variables of the causal models.
    \item $\mathcal{V}$ denotes the set of observed variables consisting of three mutually exclusive sets, 1) set of sensitive attributes $S$, 2) set of data attributes $X$ and 3) label $Y$; i.e, $\mathcal{V} = S \cup X \cup Y$.
    \item $\mathcal{F} = \{ F_1, F_2, \ldots F_n \}$ is a set of functions that define the relationship between $v_i \in \mathcal{V}$ and $Pa_i$ for all $i$. In other words, $v_i = F_i(Pa_i) + \eta_i$, where $\eta_i$ is a random variable drawn from a distribution and $Pa_{i} \subseteq \mathcal{V} \cup \mathcal{U}$. 
\end{itemize}
Defining the set of functions $\mathcal{F}$ is crucial for generating counterfactual samples of $S$ and performing related computations (for details see \citeauthor{kusner2017counterfactual}~\shortcite{kusner2017counterfactual} and \citeauthor{pearl2009causal}~\shortcite{pearl2009causal}). For intuitive examples explaining the variables $(\mathcal{U} ,\mathcal{V},\mathcal{F})$ in relation to causal models, see Section \ref{pipeline}  and Fig. \ref{fig_causal_models}.
\subsection{Counterfactual sample generation}
To create an augmented dataset including actual observations and counterfactual samples of $S$  we take the following steps (see chapter 4 of \citeauthor{pearl2009causal}{}):
\begin{itemize}
    \item We assume a prior distribution over $\mathcal{U}$ and infer its posterior distribution given $\mathcal{V}$.
    \item Intervene on sensitive attributes $S \subseteq \mathcal{V}$ to generate counterfactual samples of $S$. The counterfactual samples of $S$ are then augmented with actual observations to create augmented dataset (See Section \ref{pipeline} for details).
\end{itemize}
\subsection{Counterfactual fairness}
We adapt the definition of counterfactual fairness \cite{russell2017worlds} and use $Y_{S \rightarrow s'}$ to denote the label of the counterfactual sample of $S$. A predictor $\hat{Y}$ of $Y$ is said to be counterfactually fair given the observed data attributes $X$ and sensitive attributes $S$ if 
$P(\hat{Y}_{S\leftarrow s}=y|X=x,S=s)  
= P(\hat{Y}_{S\leftarrow s'}=y|X=x,S=s)$
for all $s' \neq s$ and all $y$.
Intuitively, this equation ensures that the prediction probability remains unaffected by interventions on sensitive attributes $S$ when all other attributes are same. For example, if we observe that talks given by white male speakers are rated to be fascinating with a  probability of 0.6 then counterfactually assigning the same talk content and other attributes to white females, say, should not change the probability from 0.6.
\section{Data}\label{data}
\begin{table}
  \caption{TED talk Dataset Properties: Information about the TED talk videos that are used in the causal DAGs}
  \begin{center}
  \begin{tabular}{|c|c|}
    \hline
    \textbf{Property}& \textbf{Quantity}\\
    \hline
    \textbf{Total number of talks} & 2,383\\
    \hline
    \textbf{Total number of views} & 4206,164,936\\
    \hline
    \textbf{Total length of all talks} & 564.63 Hours\\
    \hline
    \textbf{Total number of ratings} & 5,954,233\\
    \hline
\end{tabular}
\end{center}

  \label{tab:datasize}
\end{table}

The data analyzed in our study was collected from TED talk website (\url{ted.com}). We crawled the website to obtain data from 2400 videos published between 2006 and 2017, covering a wide range of topics such as cultural, social and scientific issues. This not only highlights the vast appeal and diversity of TED talks but also exhibits the importance of fair rating predictions. We have removed 17 talks from our dataset as those were not held at a public speaking set up.\\ 
The preliminary data contains details about the total number of \textit{views} $(V)$, the \textit{transcripts} ($T$) used by the speaker, \textit{ratings} ($Y$) of the videos given by the viewers, etc. The rating $Y$ for each video consists of 14 labels such as \textit{beautiful}, \textit{ingenious} and  \textit{confusing}. Summary of the dataset is given in Table \ref{tab:datasize}. 

\subsection{Data Annotations}\label{dataannot}
We use Amazon mechanical turk to collect data on the protected attributes $S$ (\textit{race} and \textit{gender}). Each video was annotated by $3$ turkers and we verified the inter-rater reliability using Krippendorff's alpha~\cite{krippendorff2011agreement} which gives an average agreement of \textbf{$93\%$}. The remaining data was manually investigated and annotated.

\subsection{Data Preprocessing}\label{datapreproc}
\begin{itemize}
    \item We obtain embedded transcript ($T \in \mathbb{R}^d$) using the doc2vec implementation of Gensim package \cite{le2014distributed} and use $d = 200$ for all reported results. 
    
    \item The original view count ($V_{old} \in \mathbb{Z}$) in the data ranges over large values compared to other  attributes. We use the min-max technique to normalize $V_{(old)}$ and obtain $V \in \mathbb{R}$.
$$
V = \frac{V_{(old)}-\min\{V_{(old)}\}_{talks}}{\max\{V_{(old)}\}_{talks}-\min\{V_{(old)}\}_{talks}}
$$
We assume that, how long a video has been online gets inherently captured by the total views and does not need to be explicitly modeled.
    \item We denote each original rating as $Y_{(old)} = (y_{1(old)}, \cdots,$ $ y_{14(old)} ) \in \mathbb{Z}^{14}$, where $y_{i(old)}$ is the count of $i^{th}$ label  from viewers.
    $Y_{(old)}$ is scaled w.r.t corresponding total ratings to acquire $Y \in \mathbb{R}^{14}$ as, 
$$
y_{i} = \frac{y_{i(old)}}{\sum_{j=1}^{14} y_{j(old)}}
$$
    \item   We binarize each rating label $y_j$ by thresholding w.r.t median $m_j$ ($median\{y_j\}_{talks}$). For each $j$ the label $y_j$ then becomes $0$ or $1$, where $1$ indicates the $y_j > m_j$. We train our classifier to predict these 14 binarized rating labels. The attributes of our final dataset are shown in Table \ref{tab:final_data}.

\end{itemize}

\begin{table}
  \caption{Pre-processed Dataset Attributes: Final attributes from the TED talk dataset that are used to train the FairyTED classifier}
  \begin{center}
  
  \begin{tabular}{|c|c|}
    \hline
    \textbf{Sensitive attributes}& $S$, race and gender\\
    \hline
    \textbf{Data attributes} & $X$, transcript and view count\\
    \hline
    \textbf{Label} & $Y$, rating \\
    \hline
\end{tabular}
\end{center}

  \label{tab:final_data}
\end{table}

\section{Observation in Data}\label{observation}
\begin{figure}
\centering
\includegraphics[width=0.95\columnwidth]{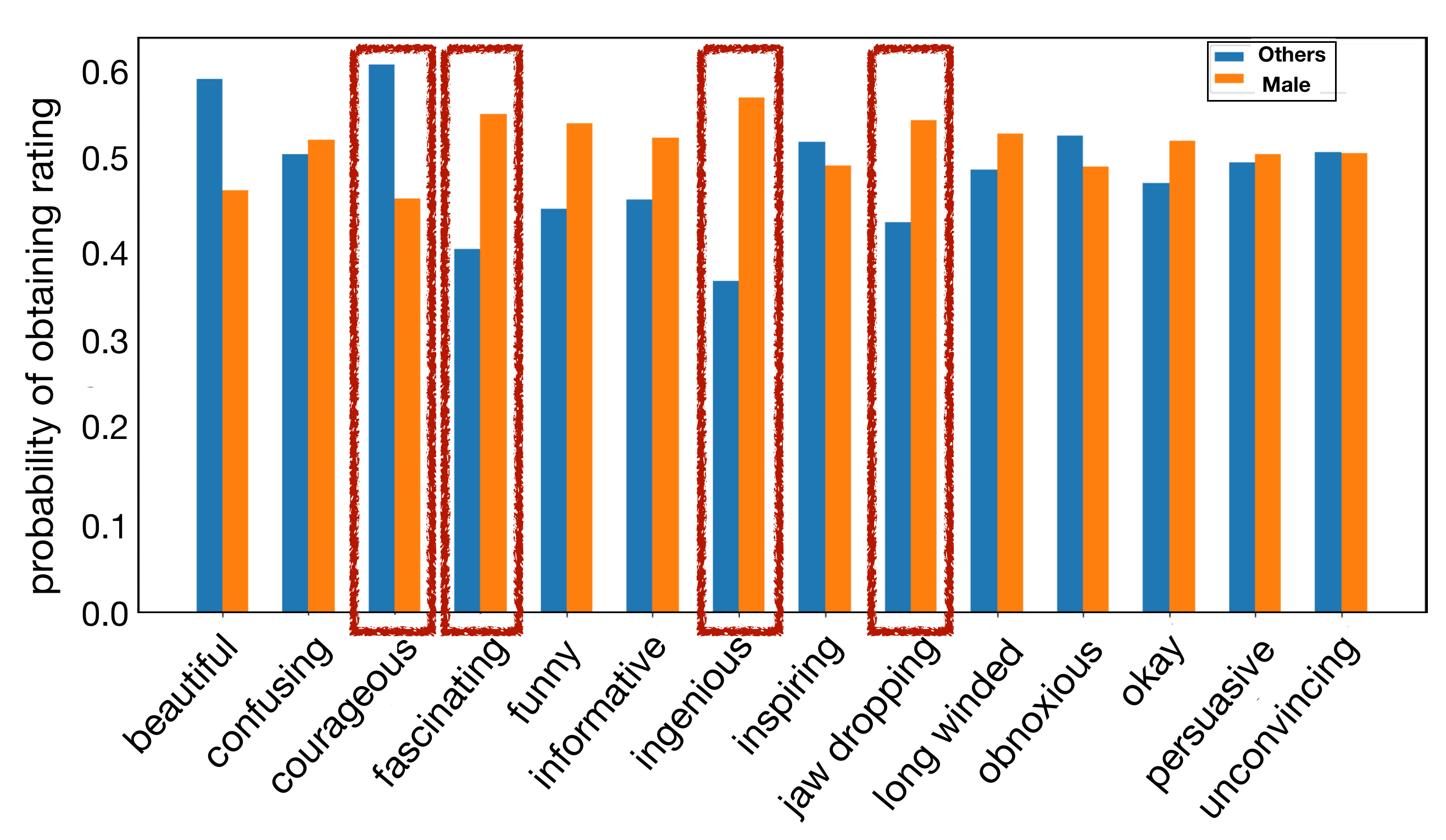} 
\caption{Existence of bias in data: Here we show an example of biased rating in data using disparate SPD and DI metrics (see Observation in Data). In this example we compare male speakers with others and find male speakers are rated to give fascinating, ingenious and jaw dropping talks with higher probability. However contrary to expectation, other speakers had higher probability of being courageous.}
\label{fig_aif360metric}
\end{figure}

We used an open source tool-kit AIF360 \cite{aif360-oct-2018} to examine existing \textit{bias} or \textit{unfairness} in our preprocessed dataset w.r.t. $S$ (race and gender of the speaker). We calculated the \textit{statistical parity difference} (SPD) and \textit{disparate impact} (DI)~\cite{biddle2006adverse} for each of the 14 binarized rating labels as,
\begin{gather*}
\textrm{SPD} = \mathbb{P}(y_i = 1 | S \in \textrm{Grp 1})-\mathbb{P}(y_i=1|S \in \textrm{Grp 2})\\
 \textrm{DI} = \mathbb{P}(y_i = 1 | S \in \textrm{Grp 1})/\mathbb{P}(y_i = 1 | S \in  \textrm{Grp 2})
\end{gather*}
Using these metrics we calculate the marginal probability of $y_i$ for each $i$  across various groups and observe many significant differences. Fig. \ref{fig_aif360metric} shows some examples where we compare \textit{male} speakers with speakers of \textit{other genders}. The difference between blue and orange bars are noticeable for rating labels marked with red blocks. We observe that talks from male speakers are rated \textit{ingenious}, \textit{fascinating} and \textit{jaw dropping} with greater probability. This identifies some classic instances of bias in data arising from social norms and structures. However, not all bias observed in data are against the presumed \textit{unprivileged} community, for example, speakers from other genders get higher probability for \textit{courageous} label as compared to male speakers. Our goal is to remove all types of bias from data, both expected and unexpected. The counterfactual fairness is agnostic to the type of bias and aims to remove all possible unfairness in rating across all possible combinations of sensitive attributes. Moreover it can be shown that under suitable assumptions, counterfactual fairness implies group fairness (see arxiv version for details).
\section{FairyTED Pipeline}\label{pipeline}
\begin{figure*}[t]
\centering
\includegraphics[width=.95\textwidth]{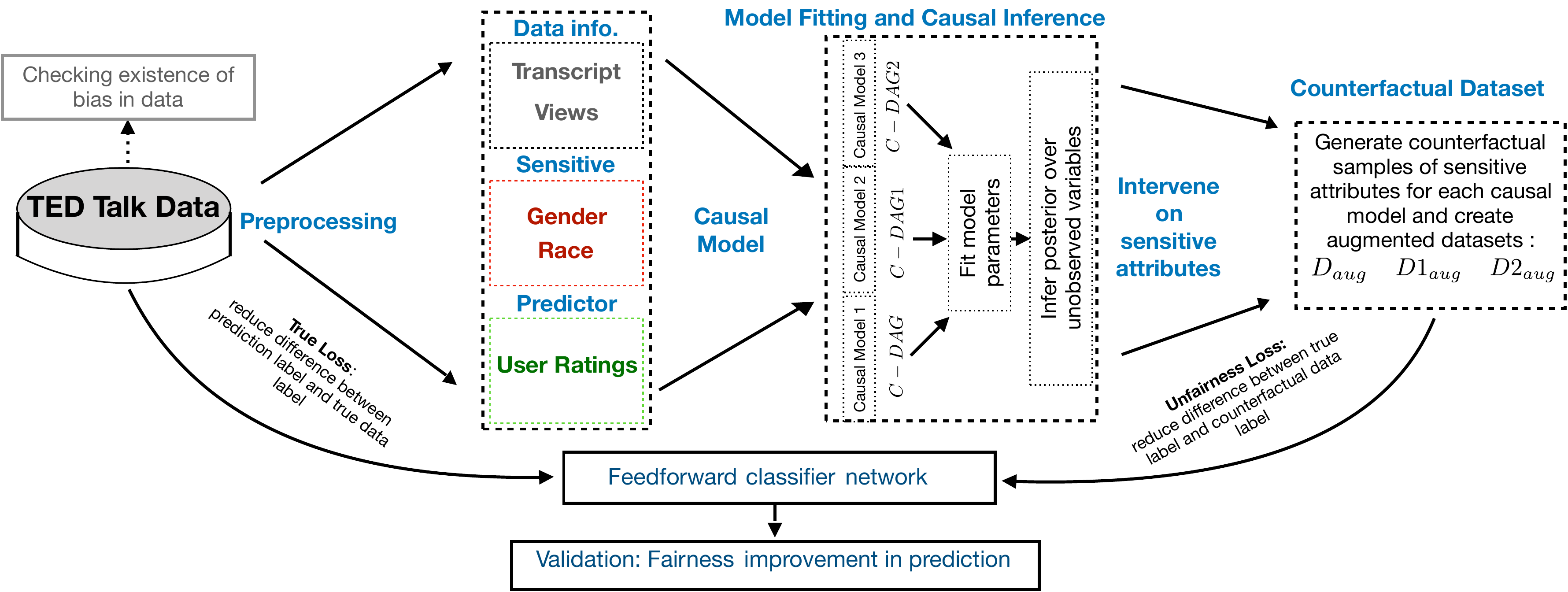} 
\caption{FairyTED pipeline: The setup to build a fair classifier for TED talk dataset (see Section \ref{pipeline}). First step is to confirm presence of bias in data. Next we preprocess the data to obtain relevant attributes (details in Section \ref{data}). Following that we build possible causal models for the preprocessed data and generate augmented datasets with counterfctual samples (see Section \ref{causalmodelsub}). We then train a classifier on the augmented datasets with a loss function incorporating a fairness measure (see Section \ref{classifier}). As a final step we validate that our system learns to generate fair prediction of ratings for the TED talk dataset (see Results).}
\label{fig_model_pipeline}
\end{figure*}

To achieve the goal of building a fair predictor of TED talk ratings $Y$, we execute the following steps:
\begin{enumerate}
    \item \textbf{Show bias in real data:} We show that the viewer ratings in the TED talks is actually biased by using SPD and DI. This justifies the need to build a fair classification model for TED talk ratings.
    \item \textbf{Preprocess data and define causal model:} We then define a causal model, C-DAG of the TED talk ratings, which consist of three major components: the unobserved variables $U$, the sensitive attributes $S$ and the TED talk video attributes $X = \{T,V,Y\}$ obtained by preprocessing the data (see Section \ref{datapreproc}). For a specific model shown in Fig. \ref{fig_causal_models},  $U$ is the skill set and background of the speaker and $S$ is gender and race. $U$ and $S$ causally influence $T$, 
    $V$ and $Y$. The skill set, background, gender and race of the speakers strongly influence their life experiences and hence govern the content of their talks. These also determine whether viewers choose to view their talk or not and what type of rating they get if viewed. Since other similar models can also be justified for the dataset, we ensure that our system is robust to any kind of causal model with similar setup (see Fig. \ref{fig_causal_models}).
    \item \textbf{Model average:} We consider variants of C-DAG with two intuitively possible manipulations, 1) C-DAG1: unobserved causes affecting $T$ and $V$ are independent, meaning $U$ decomposes into $U_1$ and $U_{2}$ such that $U_1$ affects $T$ and $U_{2}$ affects $V$ (e.g, skill set only influences how likely a talk will be viewed and background of the speaker influences the content of the talk, shown in Fig. \ref{fig_causal_models}(b)). 2) C-DAG2: the affect of sensitive attributes is manipulated, we consider the case where gender does not influence $T$ (Fig. \ref{fig_causal_models}(c)).
    \item \textbf{Fit model parameters:} We fit the parameters of each causal model, C-DAG, C-DAG1, C-DAG2. 
    \item \textbf{Create augmented data with counterfactual samples:} Next from each of these fitted models, we generate counterfactual samples of $S$ (such as replacing male speaker with female speaker for a particular talk with fixed skill set and background, also see Preliminaries) and create augmented datasets ($D_{aug}, D1_{aug}, D2_{aug}$) to be used for classification. 
    \item \textbf{Train fair classifier:} Finally for each model we use corresponding augmented dataset to train a neural network for binary classification of each of the 14 rating labels. The loss function used to train the network has two parts: 1) the first part minimizes prediction error when compared to true data labels and 2) the second part reduces disparity between the labels of observed values of $S$ and their corresponding counterfactuals. This ensures that simply changing a male speaker to female with fixed skill set and background does not influence the rating. The prediction accuracy of the fair classifier is obtained by averaging performance across all three models.
    \item \textbf{Fairness validation:} We finally validate that our classifier is counterfactually fair as compared to actual ratings provided by viewers. For this we introduce a novel metric \textit{coefficient of probability variance}, $CV_{prob}$ that compares variability of ratings across possible instances of $S$ before and after introduction of fairness measure (e.g. hypothetically if male and female speakers were rated funny with probability 0.75 and 0.45 just due to difference in $S$ and after fairness was introduced in prediction this variability dropped, becoming 0.75 and 0.70 solely based on same content, skill set and background).
\end{enumerate}
This pipeline brings together fairness measuring metrics and counterfactual fairness incorporation techniques to build a complete setup for TED talk dataset (Fig. \ref{fig_model_pipeline}). Our setup can be applied to any language or video dataset  whose feature embedding can be obtained using any state-of-the-art method. In addition, this setup can accommodate multiple causal models to ensure fairness in classifier across all possible models. Besides this, our setup also allows having nodes that cater to unobserved causes in the world.

\subsection{Causal Model}\label{causalmodelsub}
\begin{figure*}[t]
\centering
\includegraphics[width=.95\textwidth]{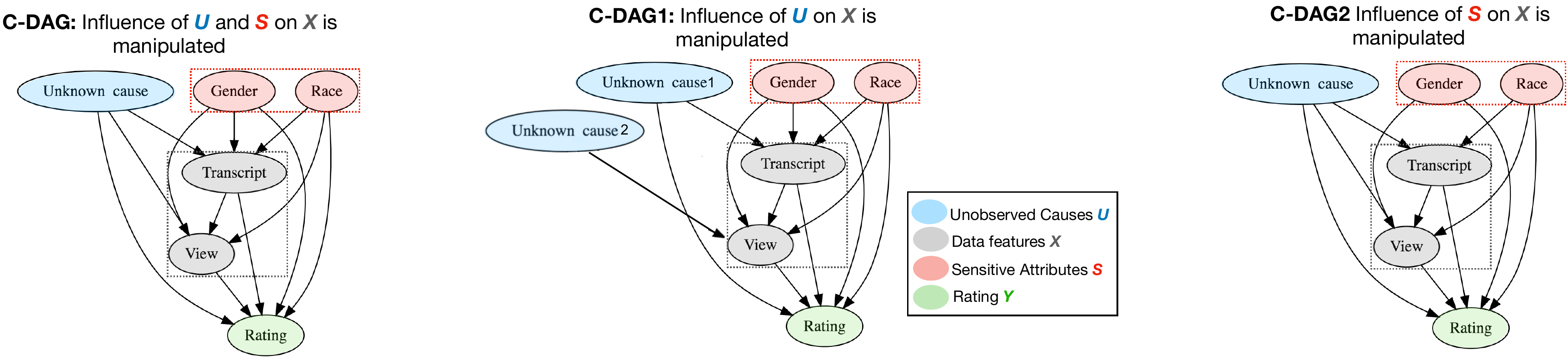} 
\caption{Causal models: Here we show three causal models considered for our FairyTED setup. C-DAG consists of all relevant dependencies between unobserved attributes $U$, sensitive attributes $S$, data attribute $V$(view) and rating prediction $Y$. In C-DAG1 we manipulate the influence of $U$ on $V$ and in C-DAG2 we manipulate the influence of $S$ on $V$. This setup allows averaging across possible causal models when the true model is unknown.}
\label{fig_causal_models}
\end{figure*}

We consider three relevant variants of causal models for TED talk data using the general definitions mentioned in preliminaries.
\begin{itemize}
    \item C-DAG as in Fig. \ref{fig_causal_models}(a),
\begin{align*}
    T & \sim \mathcal{N}\left( w_T ^U U + w_T ^S S,\sigma_T ^2 \mathbf{1} \right)\\
    V & \sim \mathcal{N}^{[0,1]}\left( \sigma \left(w_{V} ^U U + w_{V} ^S S + w_{V} ^T T\right),\sigma_{V} ^2 \mathbf{1} \right)\\
    Y & \sim Bern\left( \sigma\left(w_Y ^U U + w_Y ^S S + w_Y ^T T + w_Y ^{V} V \right)\right)
\end{align*}

where $\mathcal{N}^{[0,1]}$ denotes the truncated Gaussian in the domain $[0,1]$.
We generate $U$ from $\mathcal{N} \left(0,\mathbf{1} \right)$ and fit $\Theta_{\textrm{C-DAG}} = \{ w_T^U, w_T^S, \sigma_T, w_{V}^U, w_{V}^S, w_{V}^T,\sigma_{V}, w_Y^U, w_Y^S, w_Y^T, w_Y^{V}, \sigma_Y\}$ using variational inference algorithm in PyMC3~\cite{Salvatier2016}.
\item We modify C-DAG such that $U$ decomposes into mutually exclusive sets $U_1$ and $U_2$ to influence $T$ and $V$ independently (Fig. \ref{fig_causal_models}(b)) giving C-DAG1,
\begin{align*}
    T & \sim \mathcal{N}\left(w_T ^{U_1} U_1 + w_T ^S S,\sigma_T ^2 \mathbf{1} \right)\\
V & \sim \mathcal{N}^{[0,1]}\left(\sigma(w_{V} ^{U_2} U_2 + w_{V} ^S S + w_{V} ^T T),\sigma_{V} ^2 \mathbf{1} \right)\\
 Y & \sim Bern\left(\sigma\left(w_Y ^{U_1} U_1 + w_Y ^S S + w_Y ^T T + w_Y ^V V \right)\right)
\end{align*}
Both $U_1$ and $U_2$ are drawn from $\mathcal{N} \left(0,\mathbf{1} \right)$ and we fit $\Theta_{\textrm{C-DAG1}} = \{w_T ^{U_1}, w_T^S,\sigma_T, w_{V} ^{U_2}, w_{V} ^S, w_{V} ^T,\sigma_{V}, w_Y ^{U_1}, w_Y ^S,$ $ w_Y ^T, w_Y ^{V},\sigma_Y\}$ to obtain posterior distributions over $U_1$ and $U_2$.
\item  In C-DAG2 $S$ from C-DAG decomposes into $S_1$ and $S_2$ to influence $T$ and $V$ as in Fig. \ref{fig_causal_models}(c). $S_1 \subset S$ consists of race only, whereas, $S_2 = S$ which includes both race and gender,
\begin{align*}
    T & \sim \mathcal{N}\left(w_T ^U U + w_T ^{S_1} S_1,\sigma_T ^2 \mathbf{1} \right)\\
V & \sim \mathcal{N}^{[0,1]}\left(\sigma(w_{V} ^U U + w_{V} ^{S_2} S_2 + w_{V} ^T T),\sigma_{V} ^2 \mathbf{1} \right)\\
 Y & \sim Bern \left(\sigma\left(w_Y ^U U + w_Y ^S S + w_Y ^T T + w_Y ^{V} V \right)\right)
\end{align*}
Here, $U \sim \mathcal{N} \left(0,\mathbf{1} \right)$ and we fit $\Theta_{\textrm{C-DAG2}} = \{w_T ^{U}, w_T ^{S_1},\sigma_T, w_{V} ^{U}, w_{V} ^{S_2}, w_{V} ^T,\sigma_{V}, $
$w_Y ^{U}, w_Y ^S, w_Y ^T, w_Y ^{V},\sigma_Y\}$ to obtain posterior distribution over $U$.
\end{itemize}
From each of the three models we generate counterfactual samples of $S$.  

\subsection{Classifier Model}\label{classifier}
We train a neural network with one hidden layer of 400 nodes to predict ratings $Y$. Let $\{(s_i,x_i)\}_{i=1}^N$, $\{y^i\}_{i=1}^N$ represent the attributes and labels of the dataset where $s_i$ represents an instance of $S$ and $x_i$  represents instance of $(T, V)$. We train a classification function $g$ such that $\hat Y = g(s,x)$ using a loss function which is a combination of \textit{prediction loss} and \textit{unfairness loss}. We use binary cross entropy loss (BCE) to calculate the prediction error and an unfairness function $u$ to estimate the unfairness of the classifier as, 

\begin{equation}\label{loss_function}
\begin{split}
    \mathcal{L}(g) = \frac{1}{N} \sum_{i=1}^N \bigg(\underbrace{\textrm{BCE}\left(g(s_i,x_i), y^i\right)}_{\textit{prediction loss}}  \\+\underbrace{\gamma   \sum_{s'\neq s_i} u(g, s',s_i, x_i)}_{\textit{unfairness loss}}  \bigg)
\end{split}
\end{equation}
\begin{equation}
u(g, s',s_i, x_i) = \frac{1}{C} \sum_{c=1}^C \max\{0, |g(s_i, x_i^c)-g(s'_i, x_i^c)|-\epsilon\}
\end{equation}
where $C$ represents the number of counterfactual samples for each observed data instance and $\epsilon$ is a hyperparameter which makes sure that our predictor maintains a $(\epsilon,\delta)$- approximate counterfactual fairness ($\delta$ is a function of $\gamma$, for more details about the choice of the unfairness function, please refer to \cite{russell2017worlds}). We tune $\gamma$ and $\epsilon$ to obtain best results in our causal models, see Table 3. 

\section{Results}\label{results}
\subsection{Prediction Accuracy}
After training the classifier with augmented datasets for the three models, we obtain an average prediction accuracy of $69\%$ across all rating labels (Table \ref{accuracy}). This accuracy is obtained by training the classifier without the unfairness measure in the loss function. The mean accuracy obtained from an unfair classifier is comparable to accuracy reported in recent studies on TED talk data ~\cite{cheng2014predicting,tanveer2019causality}. However, note that our language model~\cite{le2014distributed} is much simpler when compared to methods used in the cited studies. With this simple choice we emphasize the general appeal of our approach whose goal is to reduce unfair prediction in data irrespective of the embedding technique. After addition of the unfairness measure $u$ in the loss function in equation \eqref{loss_function}, the average prediction accuracy goes down as expected, to $67\%$ but not significantly (see section \ref{classifier} for details). The hyperparameter $\gamma$ in the loss function plays a critical role in determining the trade-off between fairness in prediction and its accuracy, the smaller the value of $\gamma$, the more unfair the prediction.

\subsection{Fairness Improvement}
\begin{figure*}[t]
\centering
\includegraphics[width=.95\textwidth]{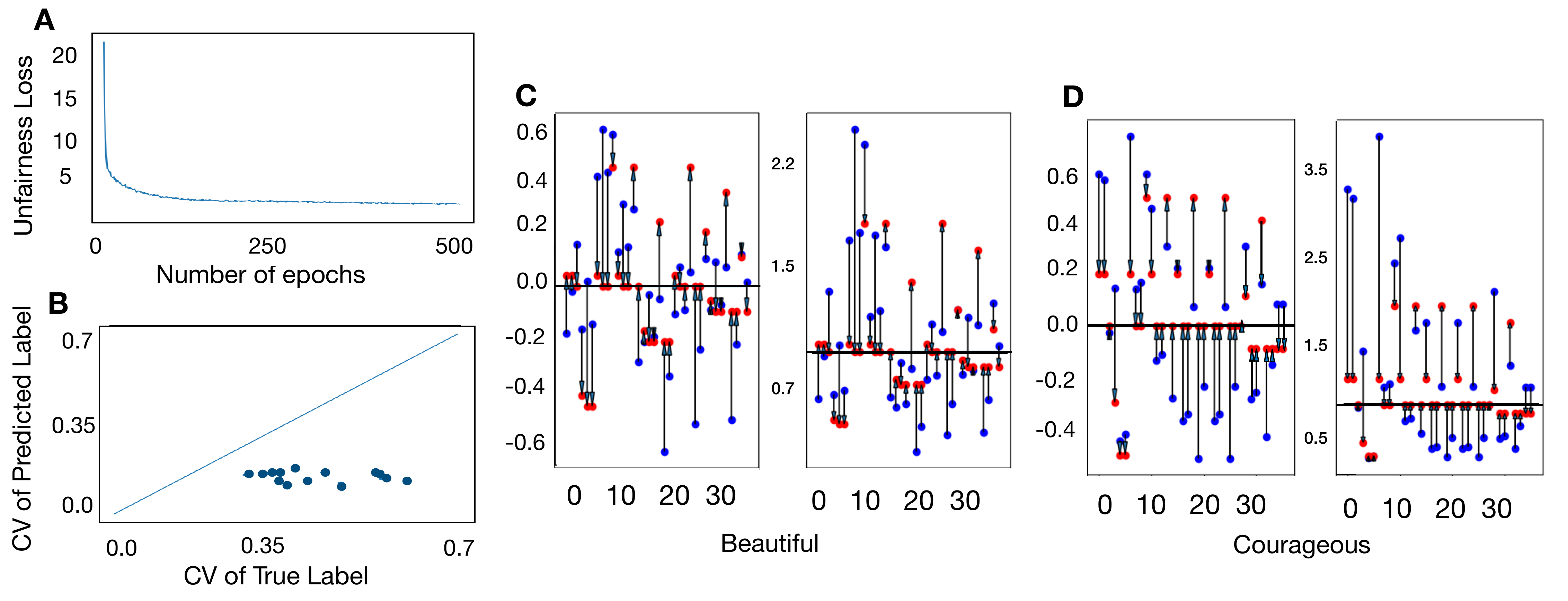} 
\caption{Increase in fairness of classifier: (A) The unfairness measure of the classifier decreases with the increase in training iterations (B)  $CV_{prob}$ values of true vs predicted labels. All points fall under the $y=x$ diagonal line, implying $CV_{prob}$ is reduced by the fair classifier. (C) SPD moves towards 0 and (D) DI moves towards 1 for most of all possible pairs of groups across $S$ after the prediction. The blue dots denote the fairness measures for the true labels and red dots denote the fairness measures for the predicted labels}
\label{result_fig}
\end{figure*}

We first show that there is a decrease in the unfairness measure of the classifier with the increase in training iterations (Fig \ref{result_fig}(A)). 
We then verify the improvement of fairness in prediction of all 14 rating labels across possible groups of $S$. To do so we come up with a fairness comparison metric $CV_{prob}$ and compare prediction fairness before and after addition of unfairness function $u$ in the loss function in equation \eqref{loss_function}. 

\begin{table}[ht]
\caption{The accuracy of the classifier is reported  for different set of hyperparameters $(\epsilon,\gamma)$.  Note that the accuracy increases if $\gamma$ is reduced or $\epsilon$ is increased (increase unfairness). When $(\epsilon,\gamma)=(\infty,0)$ (most unfair) it achieves the highest accuracy.}
\begin{center}
\resizebox{.95\columnwidth}{!}{
\label{accuracy}
\begin{tabular}{ |c|c|c|c|c| } 
\hline
 $(\epsilon, \gamma)$& (0.1,0.2)& (0.5,0.2) & 0.1,0.4)&$(\infty,0)$\\
 \hline
 \textrm{beautiful} & 0.7  & 0.7  & 0.66  & 0.77  \\
\textrm{confusing} & 0.63  & 0.62  & 0.6  & 0.66   \\
\textrm{courageous} & 0.71  & 0.72  & 0.69  & 0.74   \\
\textrm{fascinating} & 0.68  & 0.62  & 0.64  & 0.7   \\
\textrm{funny} & 0.71  & 0.67  & 0.67  & 0.71   \\
\textrm{informative} & 0.71  & 0.66  & 0.65  & 0.7   \\
\textrm{ingenious} & 0.7  & 0.65  & 0.65  & 0.7   \\
\textrm{inspiring} & 0.66  & 0.68  & 0.61  & 0.72   \\
\textrm{jaw-dropping} & 0.64  & 0.61  & 0.58  & 0.66   \\
\textrm{longwinded} & 0.64  & 0.57  & 0.6  & 0.61   \\
\textrm{obnoxious} & 0.58  & 0.59  & 0.59  & 0.62   \\
\textrm{ok} & 0.62  & 0.59  & 0.57  & 0.66   \\
\textrm{persuasive} & 0.71  & 0.67  & 0.66  & 0.75   \\
\textrm{unconvincing} & 0.62  & 0.62  & 0.63  & 0.66  \\ 
\hline
\end{tabular}
}
\end{center}
\end{table}

\subsection{$CV_{prob}$ Metric}

We have 3 types of gender (male, female and other) and 4 types of race (White, Asian, African American and other) under $S$ giving 12 possible groups denoted by $G=\left\{ G_1, G_2,\cdots, G_{12} \right\}$ on whom counterfactual fairness is tested. We then calculate with what probability each of these 12 groups obtain a particular rating, i.e., for all $k \in \{1,\cdots,14\}$ and $i \in \{1,\cdots,12\}$, we compute $p_{i}^{k}= \mathbb{P}( y_k=1| S = G_i)$. 
We denote,  $P^{k} = (p_1^{k},\ldots p_{12} ^{k})$. Similarly for the predicted label we compute $\hat{p}_{i}^{k}= \mathbb{P}(\hat{y}_k = 1| S = G_i)$ and denote,  $\hat{P}^{k} = (\hat{p}_1^{k},\ldots, \hat{p}_{12} ^{k}).$ Note that, variability in the coordinates of $P^{k}$ is a measure of fairness across the group for the rating $y_k$. In particular, the more variable these coordinates are, the more unfair the label is. Also $CV = \frac{std}{mean}$ is a common statistical metric used to quantify variability/irregularity in a set of values. Hence we define, $CV_{prob}^k$ as $CV$ for the coordinates of $P^{k}$. Similarly, $\hat{CV}_{prob}^k$ is the $CV$ for the coordinates of $\hat{P}^{k}.$ Hence, if a predictor improves fairness in the prediction, then $\hat{CV}_{prob}^k$ should be less than $CV_{prob}^k$ as in Fig. \ref{result_fig}(B). 
We also compared the SPD and DI for the true labels and predicted labels of the test dataset (Fig. \ref{result_fig} (C,D)).

\section{Conclusion}
The FairyTED setup is applicable to any dataset with the following properties, 1) contains sensitive attributes which can cause biased predictions,
2) it is possible to define a causal model with relevant attributes of the dataset,  
3) a counterfactual fairness measure can be defined on the prediction probability by using counterfactual samples of the sensitive attributes.\\
We successfully identified the above properties in the TED talk dataset and removed bias/unfairness in rating prediction. The impact of this work is many-fold: 
1) First, it identifies the necessity of applying counterfactual fairness on the rich and influential database of TED talk videos. 
2) Second, it identifies that counterfactual fairness measure is the most relevant for TED talk videos and like datasets. This is because it allows identification of attributes in data which play a critical role in causing biased predictions for example male versus female speakers. In public speaking platforms it is expensive and implausible to test the change in rating of a talk when given by a female speaker instead of a male speaker with the same content and same skill set. Counterfactual fairness allows to hypothetically test any such example and correct for the resulting bias. 
3) In most real world situations it is hard to know the true causal model for the observed dataset, and our setup can deal with this issue as long as we have some idea of possible models.
4) In rich and complicated domain such as TED talks, there can be a lot of unobserved attributes that can affect the data and our setup can take care of it by inferring a posterior distribution over the unobserved attributes.
5) Any dataset with a simple embedding scheme can use this model. One straightforward extension of this framework can be on job interview based datasets. We can include better encoding schemes such as~\cite{devlin2018bert,yang2019xlnet} to obtain even better prediction accuracy besides making it fair. We can also include multi-modal (e.g. audio-visual) information from the dataset to obtain rich representation and test how counterfactual fairness measure generalizes across various modes. 
6) We propose an intuitive novel metric that quantifies degree of fairness employed by our setup.
Besides these, information from temporal evolution can also be used to improve our framework in future work. Finally the FairyTED setup has three important social impacts: 1) It ensures that speakers get fair feedback and can improve only based on fair fallacies 2) Organizers can employ diverse speakers without worrying about degradation in rating when skill, ability, influence and content are matched. For example, suppose a talk on global warming was given by a male speaker and obtained good rating. With our setup, the organisers can choose a female speaker with comparable qualities as the male to speak on the global warming without worrying about deterioration in rating. 3) Finally, new viewers will not be biased by prevalent unwanted biased ratings from previous users and prevents propagation of unfairness over time.
\section{ Acknowledgments}
We thank the anonymous reviewers for their valuable suggestions.

\bibliographystyle{aaai}
\bibliography{references}

\appendix
\section{Relation between Counterfactual Fairness and Group Fairness}
Let $R$ be the set of all races and $G$ be the set of all genders. Usually group fairness considers fairness between  racial groups(e.g. White vs African American) or gender groups (e.g. Male vs Female), whereas counterfactual fairness measures fairness among finer groups (e.g.  white male vs African American female). More formally, counterfactual fairness measures fairness among all (race,gender) pairs. Let,
$$pr(r,g)=\mathbb{P}(\hat{Y}=y| X = x, \text{Race} = r,\text{Gender} = g) $$
Under the assumption that counterfactual fairness tries to minimize the absolute difference between $pr(r,g)$ and $pr(r',g')$ $\forall r,r' \in R $ and $g,g' \in G$. In an ideal scenario this difference is $0$ and we call this predictor ideally and counterfactually fair. 
\begin{theorem}\label{cf_implication}
With causal model assumption (C-DAG,C-DAG1,C-DAG2) any ideally and counterfactually fair predictor satisfies group fairness w.r.t. each race group and each gender group, i.e.,we have,
$$\mathbb{P}(\hat{Y}=y| X = x, \text{Race} = r)= \mathbb{P}(\hat{Y}=y| X = x, \text{Race}= r')$$
$\forall r,r'\in R$ and
$$\mathbb{P}(\hat{Y}=y| X = x, \text{Gender} = g)$$ 
$$= \mathbb{P}(\hat{Y}=y| X = x, \text{Gender}= g')$$
$\forall g,g'\in G.$
\begin{proof}(Proof of Theorem \ref{cf_implication}]
\begin{align}
    & \mathbb{P}(\hat{Y}=y| X = x, \text{Race} = r)\nonumber &\\
    & = \sum_{g} \mathbb{P}(\hat{Y}=y, \text{Gender} = g| X = x, \text{Race} = r)&\nonumber\\
    & =\sum_{g} \mathbb{P}(\hat{Y}=y| X = x, \text{Race} = r, \text{Gender} = g)\cdot &\nonumber\\ 
    &\quad \quad\mathbb{P}( \text{Gender} = g | \text{Race} = r)&\label{gender_race_ind}\\
    & = \sum_{g} \mathbb{P}(\hat{Y}=y| X = x, \text{Race} = r, \text{Gender} = g)\cdot &\nonumber\\ 
    &\quad \quad\mathbb{P}(\text{Gender} = g )\label{cf_def}&\\
    & = \mathbb{P}(\hat{Y}=y| X = x, \text{Race} = r')&
\end{align}
\eqref{gender_race_ind} follows from the independence assumption between race and gender, \eqref{cf_def} follows from the definition of ideal counterfactually fair model.
\end{proof}
\end{theorem}
This establishes the fact that counterfactual fairness is a stronger notion than group fairness. 
\end{document}